\documentclass[a4paper]{article}

\usepackage{INTERSPEECH2019}
\usepackage{url}
\usepackage{color}
\usepackage{comment}

\title{Token-Level Ensemble Distillation for Grapheme-to-Phoneme Conversion}

\name{Hao Sun$^1$\thanks{This work was done while the first author was an intern at Microsoft.}, Xu Tan$^2$, Jun-Wei Gan$^3$, Hongzhi Liu$^1$, Sheng Zhao$^3$, Tao Qin$^2$ and Tie-Yan Liu$^2$}

\address{
  $^1$Peking University\\
  $^2$Microsoft Research\\
  $^3$Microsoft STC Asia}
\email{sigmeta@pku.edu.cn, xuta@microsoft.com, junwg@microsoft.com, liuhz@pku.edu.cn, Sheng.Zhao@microsoft.com, taoqin@microsoft.com, tyliu@microsoft.com}

\begin{document}

\maketitle
\begin{abstract}

Grapheme-to-phoneme (G2P) conversion is an important task in automatic speech recognition and text-to-speech systems. Recently, G2P conversion is viewed as a sequence to sequence task and modeled by RNN or CNN based encoder-decoder framework. However, previous works do not consider the practical issues when deploying G2P model in the production system, such as how to leverage additional unlabeled data to boost the accuracy, as well as reduce model size for online deployment. In this work, we propose token-level ensemble distillation for G2P conversion, which can (1) boost the accuracy by distilling the knowledge from additional unlabeled data, and (2) reduce the model size but maintain the high accuracy, both of which are very practical and helpful in the online production system. We use token-level knowledge distillation, which results in better accuracy than the sequence-level counterpart. What is more, we adopt the Transformer instead of RNN or CNN based models to further boost the accuracy of G2P conversion. Experiments on the publicly available CMUDict dataset and an internal English dataset demonstrate the effectiveness of our proposed method. Particularly, our method achieves 19.88\% WER on CMUDict dataset, outperforming the previous works by more than 4.22\% WER, and setting the new state-of-the-art results.

\end{abstract}

\section{Introduction}

Grapheme-to-phoneme (G2P) conversion aims to generate a sequence of pronunciation symbols (phonemes) given a sequence of letters (graphemes), which is an important component in automatic speech recognition and text-to-speech systems~\cite{ren2019almost,ren2019fastspeech} to provide accurate pronunciations for the words not covered by the lexicon. G2P conversion can be viewed as a sequence to sequence task and modeled by the encoder-decoder framework. ~\cite{yao2015sequence} adopt LSTM for G2P conversion and achieve improvements than the previous joint n-gram model~\cite{bisani2008joint}. ~\cite{chae2018convolutional} use convolutional sequence to sequence model and non-sequential decoding, and attain the previous best results on the public CMUDict dataset.


While previous works introduced the neural sequence to sequence models into G2P conversion and indeed achieved improvements over conventional methods, they did not take into account several practical issues of G2P conversion in the production system. First, considering training data is always costly through human labeling, how to further leverage the unlimited amount of unlabeled data is critical to improve the performance of G2P conversion. Second, large or ensemble models are too costly to serve when deploying in the online systems. How to reduce the model size but maintain high accuracy is essential.

Inspired by the knowledge distillation in computer vision~\cite{hinton2015distilling,furlanello2018born} and natural language processing~\cite{kim2016sequence,freitag2017ensemble,tan2018multilingual}, in this work, we propose the token-level ensemble distillation for G2P conversion, to address the practical problems mentioned above. First, we use knowledge distillation to leverage the large amount of unlabeled words. Specifically, we train a teacher model to generate the phoneme sequence as well as its probability distribution given unlabeled grapheme sequence, and regard the unlabeled grapheme sequence and the generated phoneme sequence as pseudo labeled data, and add them into the original training data. Second, we train a variety of models (CNN, RNN and Transformer) for ensemble to get higher accuracy, and transfer the knowledge of the ensemble models to a light-weight model that is suitable for online deployment, again by knowledge distillation. Besides, we adopt Transformer~\cite{vaswani2017attention} instead of RNN or CNN as the basic encoder-decoder model structure, since it demonstrates advantages in a variety of sequence to sequence tasks, such as neural machine translation~\cite{vaswani2017attention}, text summarization~\cite{gong2018sentence}, automatic speech recognition~\cite{zhou2018comparison}. 

We conduct experiments on CMUDict 0.7b and our internal dataset, and also leverage additional unlabeled words crawled from the web. Our proposed method significantly boosts the accuracy of G2P conversion by 4.22\% WER compared with the previous works. Specifically, Transformer model achieves higher accuracy than RNN and CNN based models, and token-level distillation outperforms sequence-level distillation. 

Our contributions are listed as follows: (1) We propose token-level ensemble distillation for grapheme-to-phoneme conversion. (2) We are the first to use unlabeled words to boost the accuracy of grapheme-to-phoneme conversion, and  also the first to introduce Transformer into this task and achieve better performance. (3) Our method achieves the state-of-the-art accuracy on CMUdict dataset, outperforming the previous best result by 4.22\% WER.

\section{Background}
In this section, we briefly review the background of grapheme-to-phoneme conversion, Transformer model, as well as knowledge distillation.

\subsection{Grapheme-to-Phoneme conversion}
The G2P conversion is the process that generating the phoneme sequence (pronunciation) according to the grapheme sequence (word). G2P conversion is necessary and important as lexicon cannot cover all words, due to many words are long-tailed and a lot of new words and compound words appear. The spelling and pronunciation are not exactly corresponding for some languages, e.g. English. What is more, the alignments between graphemes and phonemes are complex. A grapheme may correspond to no phoneme, a single phoneme or many phonemes, as shown in Table~\ref{tab:g2p}, which makes G2P a hard task.

\begin{table}[th]
  \caption{An example of the alignments between graphemes and phonemes.}
  \label{tab:g2p}
  \centering
  \begin{tabular}{ c|cccccc }
   \toprule
    $graphemes$                       & B &U&B&B&L&E            \\
    $phonemes$                       & B  & AH&null&B&AH:L&null              \\
   \bottomrule
  \end{tabular}
  
\end{table}

Joint sequence n-gram models have been widely used~\cite{bisani2008joint,chen2003conditional,wu2014encoding} for G2P conversion. Recently, sequence to sequence models have achieved great success in machine translation task~\cite{bahdanau2014neural,luong2015effective,wu2016google,hassan2018achieving}, and are soon applied on G2P conversion. ~\cite{yao2015sequence} demonstrated that sequence to sequence models outperform joint sequence n-gram models. ~\cite{toshniwal2016jointly,milde2017multitask} combined joint n-gram models with Bi-LSTM models and achieved good performance in G2P conversion. \cite{chae2018convolutional} adopted convolutional sequence to sequence model and proposed the non-sequential decoding~\cite{guo2018non} for G2P conversion, which achieved the previous state-of-the-art result on the public CMUDict 0.7b dataset.  

While these sequence to sequence models achieve good performance on G2P conversion, there is still a gap when deploying online. In this work, we propose token-level ensemble distillation based on Transformer model, which can not only boost the accuracy of the G2P conversion with unlabeled words, but also reduce the model size for online deployment.

\subsection{Transformer}

Transformer~\cite{vaswani2017attention} has achieved the state-of-the-art performance in many NLP tasks~\cite{devlin2018bert,dehghani2018universal,wangperawong2018attending,so2019evolved}. The encoder and decoder in Transformer has $N$ identical layers, and each layer in encoder consists of two different sub-layers: multi-head self-attention and feed-forward network, while the decoder has an additional multi-head attention sub-layer. Multi-head attention is to perform the attention function $h$ times in parallel, allowing the model to jointly attend to information from different representation subspaces at different positions. Residual connection is employed between each sub-layer. Transformer can better model the interactions between any two tokens in the sequence and the computation of each token in the encoder and decoder can be parallel during training, which shows advantages over the RNN based models. To the best of our knowledge, this is the first work to apply Transformer in G2P conversion.

\subsection{Knowledge Distillation}
Knowledge distillation was first introduced by~\cite{buciluǎ2006model}  for model compression, where a light student model can approximate the accuracy of a heavy and cumbersome teacher model. \cite{hinton2015distilling} first applied knowledge distillation on neural networks, and then a lot of works expand the usage of knowledge distillation to a variety of tasks, such as image classification~\cite{furlanello2018born,anil2018large,yang2018knowledge} and natural language processing~\cite{kim2016sequence,freitag2017ensemble,tan2018multilingual}. In this work, we leverage knowledge distillation to distill the knowledge from additional unlabeled word, as well as from the ensemble models, both of which are beneficial for the online production system.  

\section{Token-Level Ensemble Distillation}
In this section, we propose the token-level ensemble knowledge distillation to boost the accuracy of G2P conversion, as well as reduce the model size for online deployment.

\subsection{Token-Level Knowledge Distillation}


Denote $D=\{(x,y) \in \mathcal{X \times Y}\}$ as the training corpus which consists of the paired grapheme and phoneme sequence. A G2P model based on sequence to sequence learning aims to minimize the negative log-likelihood loss on corpus $D$:
\begin{equation}
  \mathcal{L}_{NLL}(\theta)=-\sum_{(x,y) \in D} \log P(y|x;\theta),
  \label{eq1}
\end{equation}
where the likelihood $P(y\mid x;\theta)$ can be factored by the chain-rule and formulated as the cross-entropy between the one-hot label and per-token probability:  
\begin{equation}
  \log P(y | x; \theta)=\sum ^{T_y}_{t=1} \sum ^{| \mathcal{V}|} _{k=1} \mathbf{1} \{y_t =k\}\log P(y_t =k |y_{<t},x;\theta ),
  \label{eq2}
\end{equation}
where $T_y$ is the length of the target sequence, $|\mathcal{V}|$ is the vocabulary size of the phonemes, $y_t$ is the $t$-th target token in the phoneme sequence, and $\mathbf{1}\{\cdot \}$ is the indicator function indicating the id of the phoneme in vocabulary.

In token-level knowledge distillation, the one-hot label becomes the probability distribution output of the teacher model:
\begin{equation}
\begin{split}
  \mathcal{L}_{KD}(\theta)=-\sum_{(x,y) \in D} \sum^{T_y}_{t=1} \sum^{|\mathcal{V}|}_{k=1} Q(y_t=k| y_{<t},x;\theta_T) \\
  \times \log P(y_t=k|y_{<t},x;\theta),
  \label{eq4}
  \end{split}
\end{equation}
where $Q(y_t=k|y_{<t},x;\theta_T) $ is the probability distribution output of the teacher model $\theta _T$. 

\subsection{Ensemble Distillation with Diverse Models}
Model ensemble can incorporate the advantages of individual models, and reduce the effect of overfitting in a spirit of the bagging method~\cite{dietterich2000ensemble}. However, the online production system cannot support large ensemble models for G2P conversion. Knowledge distillation is an effective way to distill the knowledge from strong ensemble models into single model. The ensemble distillation can be formulated as follows:
\begin{equation}
\begin{split}
  \mathcal{L}_{KD}(\theta)=-\sum_{(x,y) \in D} \sum^{T_y}_{t=1} \sum^{|\mathcal{V}|}_{k=1} \bar{Q}(y_t=k|y_{<t},x) \\
  \times \log P(y_t=k|y_{<t},x;\theta),
  \label{eq6}
  \end{split}
\end{equation}

\begin{equation}
\begin{split}
 \bar{Q}(y_t=k|y_{<t},x) =\frac{\sum _{m=1}^{M} Q(y_t=k|y_{<t},x;\theta^m_T) }{M},
  \label{eq7}
  \end{split}
\end{equation}
where $\bar{Q}$ is the probability distribution combined by $M$ models ($\theta^1_T$ to $\theta^m_T$), which is simply the average of the probability distribution of $M$ models at each step of the target sequence.
 

The performance of the individual models and the diversity between them are essential for ensemble. On the one hand, we train deeper models to achieve higher accuracy. On the other hand, we choose Transformer~\cite{vaswani2017attention}, Bi-LSTM~\cite{wu2016google}, and convolutional sequence to sequence~\cite{gehring2017convolutional} models to increase the diversity of ensemble models.

\subsection{Knowledge Distillation with Unlabeled Source Words}
In G2P conversion, it is easy to obtain abundant unlabeled source words (graphemes) from lexicon corpus of news or wikipedia. Knowledge distillation gives a way of using unlabeled source data. The teacher model can generate the target phoneme sequence given the unlabeled source grapheme sequence, and the generated phoneme sequence can be used as the label for student model. What is more, more unlabeled data can help distill the knowledge of the teacher model to the student model. In this work, we also use token-level knowledge distillation for unlabeled source words. Denote $D'=\{x \in \mathcal{X}\}$ as the corpus of unlabeled source words. The knowledge distillation loss with unlabeled source words is as follows:
\begin{equation}
\begin{split}
  \mathcal{L'}_{KD}(\theta)=-\sum_{x \in D'} \sum^{T_{y'}}_{t=1} \sum^{|\mathcal{V}|}_{k=1} \bar{Q}(y'_t=k|y'_{<t},x) \\
 \times \log P(y'_t=k|y'_{<t},x;\theta),
  \label{eq8}
  \end{split}
\end{equation}
\begin{equation}
\begin{split}
  y' \sim \bar{Q}(y|x)
    \label{eq9}，
  \end{split}
\end{equation}
where $y'$ is generated by the ensemble model (Equation~\ref{eq9}), $Q(y'_t=k|y'_{<t},x) $ is the probability distribution output of the ensemble model and is calculated by Equation~\ref{eq7}.

The total loss of our method is the weighted combination of the original negative log-likelihood loss and the knowledge distillation loss~\cite{kim2016sequence,tan2018multilingual} on the labeled data, as well as the knowledge distillation loss on the unlabeled data:
\begin{equation}
  \mathcal{L}_{TOTAL}(\theta)=(1-\lambda)\mathcal{L}_{NLL}(\theta)+\lambda\mathcal{L}_{KD}(\theta) + \mathcal{L'}_{KD}(\theta),
  \label{eq5}
\end{equation}
where each loss term is formulated in Equation~\ref{eq1},~\ref{eq6} and~\ref{eq8}, $\lambda$ is the weight to trade off between the two loss terms on labeled data.

\section{Experiments and Results}
In this section, we conduct experiments to verify the effectiveness of the proposed method. We first introduce the datasets used, and then describe the implementation details. At last, we report the results of our method and conduct some comparisons and analyses. 
\subsection{Experimental Setup}
\subsubsection{Datasets}
We use two datasets to evaluate our proposed method: the first one is the publicly available CMUDict 0.7b  and the other one is our internal dataset. For the public CMUDict 0.7b dataset, we use the same training/validation/test split (108952 training words, 5447 validation words and 12855 test words) as in \cite{milde2017multitask}, which is released in the CNTK toolkit\footnote{https://github.com/Microsoft/CNTK/tree/master/Examples/Sequen ceToSequence/CMUDict/Data}. The sizes of the grapheme and phoneme vocabulary are 27 and 39 respectively. To be consistent with the previous works~\cite{bisani2008joint,chae2018convolutional,milde2017multitask}, stress markings are removed and the multiple pronunciations are retained. Our internal dataset contains 184243 training words, 10837 validation words, 21678 test words, which includes uppercase and lowercase letters and stress markings. We keep the stress markings in training and ignore the stress during test. The sizes of the grapheme and phoneme vocabulary in our internal dataset are 54 and 73 respectively. We train our models on the training set and select the best hyperparameters according to the validation set.

We crawl nearly 2,000,000 unlabeled source words from the lexicon corpus of Google news\footnote{https://github.com/mmihaltz/word2vec-GoogleNews-vectors}. As the crawled data contains words of other languages, unknown tokens and spelling errors, we first filter the data by removing the words with unknown tokens and then choose the top 300,000 unlabeled words according to their similarity to the training data\footnote{We use the distance between the 1/2/3-gram distribution of training words and unlabeled words, where the 1/2/3-gram means 1/2/3 consecutive characters.}.

\subsubsection{Model Configurations}
\textbf{Ensemble Model} We train the sequence to sequence based G2P models with different model structures for ensemble, including Transformer~\cite{vaswani2017attention}, Bi-LSTM~\cite{wu2016google} and CNN based sequence to sequence model~\cite{gehring2017convolutional}. We use 4 Transformer models, 3 CNN models and 3 Bi-LSTM models with different hyperparameters for ensemble, which give the best performance on the validation set. The 4 Transformer models share the same hidden size (256) but vary in the number of the encoder-decoder layers (6-6, 6-4, 8-6, 8-4). For the 3 CNN models, they share the same hidden size (256) but vary in the number of encoder-decoder layers (10-10, 10-10, 8-8) and convolutional kernel widths (3, 2, 2) respectively. For the 3 Bi-LSTM models, they share the same number of encoder-decoder layers (1-1), but with different hidden sizes (256, 384 and 512).

\textbf{Student Model} We choose Transformer as the student model and use the default configurations (256 hidden size and 6-6 layers of encoder-decoder) unless otherwise stated. We also vary the number of layers for the encoder and decoder to analyze and compare the accuracy and memory/time cost, which is essential for online deployment.

\subsubsection{Training and Evaluation}
We implement experiments with the fairseq-py\footnote{https://github.com/pytorch/fairseq} library in PyTorch. We use Adam optimizer for all models and follow the learning rate schedule in~\cite{vaswani2017attention}. The dropout is 0.3 for Bi-LSTM and CNN models, while the residual dropout, attention dropout and ReLU dropout for Transformer models is 0.2, 0.4, 0.4 respectively. We set the $\lambda$ in Equation~\ref{eq5} to 0.9 according to the validation performance. We train each model on 8 NVIDIA M40 GPUs. Each GPU contains roughly 4000 tokens in one mini-batch. We use beam search during inference and set beam size to 10. We use WER (word error rate) and PER (phoneme error rate) to measure the accuracy of G2P conversion. Edit distance is used in PER calculation. In WER calculation, considering the multiple pronunciations, word error is counted only when the output differs from all the references, following \cite{bisani2008joint,chae2018convolutional,milde2017multitask,rao2015grapheme}.


\subsection{Results and Analyses}
\subsubsection{Achieving State-Of-The-Art Accuracy}
We first compare our method with previous works~\cite{bisani2008joint,chae2018convolutional,milde2017multitask} on CMUDict 0.7b dataset, as shown in Table~\ref{tab:result1}. Sequitur G2P \cite{bisani2008joint} is a well established G2P conversion tool using joint sequence modelling and is widely used as a baseline for comparison. \cite{milde2017multitask} used the ensemble of Bi-LSTM and joint n-gram model. The convolutional sequence to sequence model with non-sequential greedy decoding (NSGD) \cite{chae2018convolutional} is the previous state-of-the-art on CMUDict 0.7b dataset\footnote{They use a training/validation/test split different from~\cite{milde2017multitask} and ours. Therefore, we reproduce their work with on our training/ validation/test split, based on their public codebase (https://github. com/ctr4si/NSGD\_G2P), and get similar result as theirs.}. It can be seen that our method on 6-layer encoder and 6-layer decoder Transformer achieves the new state-of-the-art result of 19.88\% WER, outperforming NSGD by 4.22\% WER. 


\begin{table}[th]
  \caption{Comparison between our method and the previous works on CMUDict 0.7b dataset.}
  \label{tab:result1}
  \centering
  \begin{tabular}{ccc}
   \toprule
    \textbf{Method}                     & \textbf{PER} &\textbf{WER}          \\
    \midrule
     Sequitur G2P~\cite{bisani2008joint}                   &6.12\%&25.71\%           \\
     Bi-LSTM + n-gram~\cite{milde2017multitask}                     &5.76\%&24.88\%           \\
     CNN with NSGD~\cite{chae2018convolutional}                     &5.58\%&24.10\%          \\
      \midrule
      Our method &\textbf{4.60\%}&\textbf{19.88\%}            \\
    \bottomrule
  \end{tabular}
  \end{table}
\vspace{-0.4cm}
\subsubsection{Reducing Model Size by 6x}
Our method can also greatly reduce the model size for online deployment. We compare the WER, the number of parameters, and the inference speed between the baseline and our method, as shown in Table~\ref{tab:result3}. The baseline method just uses transformer model (6-6 layers of encoder-decoder) without leveraging the ensemble knowledge distillation and unlabeled source words. To compare the inference speed, we use the time consumed by generating the outputs of the test set (12855 words) on a single M40 GPU with 12000 max tokens in one mini-batch. It can be seen from Table~\ref{tab:result3} that our method can still reach high accuracy with 1-1 layer of encoder-decoder, which can significantly reduce the model size by nearly 6 times and the time cost by nearly 4 times compared with the baseline model, but still achieving higher accuracy in terms of WER. The reduction in model size and inference time cost demonstrate the effectiveness of our method for online deployment.

\begin{table}[th]

  \caption{Comparison of WER, number of parameters and inference time between the baseline and our method.}
  \label{tab:result3}
  \centering
  \begin{tabular}{ccccc}
   \toprule
 \textbf{Method} & \textbf{Layers} & \textbf{WER} &\textbf{Parameters}   &\textbf{Time}          \\
    \midrule
    Baseline &  6-6  &  21.07\%  &  11.09 millions  &  17.8s           \\
   Our method &  1-1  &  20.25\%  &  1.85 millions  &  4.4s    \\ 
    \bottomrule
  \end{tabular}
\end{table}
\vspace{-0.4cm}

\subsubsection{Analyses of Our Method }
We first study the effect of distilling from unlabeled source words, as shown in Table~\ref{tab:result2}. It can be seen that
unlabeled source words can boost the accuracy by nearly 1\% WER, demonstrating the effectiveness by introducing abundant unlabeled data into knowledge distillation.
\vspace{-0.1cm}
\begin{table}[th]
  \caption{Comparison of our method with and without unlabeled source words.}
  \label{tab:result2}
  \centering
  \begin{tabular}{ccc}
   \toprule
    \textbf{Method}                     & \textbf{PER} &\textbf{WER}          \\
    \midrule
      Without unlabeled data      &4.78\%&20.71\%           \\
      With unlabeled data   &4.60\%&19.88\%            \\
    \bottomrule
  \end{tabular}
\end{table}
\vspace{-0.1cm}
We also compare token-level distillation with sequence-level distillation, where the student models are directly trained on the top-1 beam search results of the teacher network. As shown in Table~\ref{tab:token}, the result demonstrate the advantage of token-level distillation.
\vspace{-0.1cm}
\begin{table}[th]
  \caption{Comparison between token-level and sequence-level distillation.}
  \label{tab:token}
  \centering
  \begin{tabular}{ccc}
   \toprule
    \textbf{Method}                     & \textbf{PER} &\textbf{WER}          \\
    \midrule
      Sequence-level      &4.71\%&20.32\%           \\
      Token-level  &4.60\%&19.88\%            \\
    \bottomrule
  \end{tabular}
\end{table}
\vspace{-0.3cm}
Furthermore, we study the effect of ensemble teacher model in knowledge distillation. As shown in Table~\ref{tab:result-ensemble}, the ensemble teacher model can boost the accuracy by more than 1\% WER, compared with the single teacher model (a Transformer model with 6-layer encoder and 6-layer decoder), which demonstrates the strong ensemble teacher model is essential to guarantee the performance of student model in knowledge distillation.
\vspace{-0.1cm}
\begin{table}[!htb]
  \caption{Comparison of different teacher models for knowledge distillation.}
  \label{tab:result-ensemble}
  \centering
  \begin{tabular}{ccc}
   \toprule
    \textbf{Method}                     & \textbf{PER} &\textbf{WER}          \\
    \midrule
      Single teacher model      &4.93\%&21.05\%        \\
      Ensemble teacher model   &4.60\%&19.88\%            \\
    \bottomrule
  \end{tabular}
\end{table}
\vspace{-0.3cm}
At last, we compare Transformer with RNN~\cite{milde2017multitask} and CNN~\cite{chae2018convolutional} based models, without using knowledge distillation and unlabeled data, as shown in Table~\ref{tab:transformer}. We can see that Transformer model outperforms the RNN and CNN based models used in previous works, demonstrating the advantage of Transformer model.
\vspace{-0.1cm}
\begin{table}[th]
  \caption{Comparison of Transformer, LSTM and CNN.}
  \label{tab:transformer}
  \centering
  \begin{tabular}{ccc}
   \toprule
    \textbf{Method}                     & \textbf{PER} &\textbf{WER}          \\
    \midrule
     Bi-LSTM + n-gram~\cite{milde2017multitask}    &5.76\%&24.88\%           \\
     CNN with NSGD~\cite{chae2018convolutional}                     &5.58\%&24.10\%          \\
      Transformer                     &4.96\%&21.07\%           \\
    \bottomrule
  \end{tabular}
  \end{table}
\vspace{-0.5cm}

\subsubsection{Results on Our Internal Dataset}

We compare our method with the previous state-of-the-art CNN with NSGD~\cite{chae2018convolutional} (which is reproduced by ourself) on our internal dataset, as shown in Table~\ref{tab:result4}. Our method outperforms CNN with NSGD by 3.52\% WER, which demonstrates the effectiveness of our method for G2P conversion.
\vspace{-0.1cm}
\begin{table}[!htb]
  \caption{Results on our internal dataset.}
  \label{tab:result4}
  \centering
  \begin{tabular}{cccc}
   \toprule
    \textbf{Method}                     & \textbf{PER} &\textbf{WER}          \\
    \midrule
      CNN with NSGD~\cite{chae2018convolutional}               &3.79\%&22.39\%           \\
      Our method  &\textbf{3.04\%}&\textbf{18.87\%}            \\
    \bottomrule
  \end{tabular}
\end{table}
\vspace{-0.5cm}

\section{Conclusion}

In this work, we have proposed the token-level ensemble distillation with unlabeled source words for G2P conversion. Experiments on the publicly available CMUDict 0.7b dataset and our internal dataset demonstrate the effectiveness of our method on both improving the accuracy of G2P conversion and reducing the model size for online deployment. For future work, we will leverage more unlabeled data and pre-training~\cite{song2019mass} to improve the performance, and extend our work to other languages. 


\bibliographystyle{IEEEtran}

\bibliography{mybib}

\end{document}